\pdfoutput=1

\documentclass[11pt]{article}

\usepackage{emnlp}

\usepackage{times}
\usepackage{latexsym}

\usepackage[T1]{fontenc}

\usepackage[utf8]{inputenc}

\usepackage{microtype}

\usepackage{soul}
\usepackage{url}
\usepackage{graphicx}
\usepackage{amsmath}
\usepackage{amsthm}
\usepackage{booktabs}
\usepackage{algorithm}
\usepackage{algorithmic}

\usepackage{amssymb}
\usepackage{caption}
\usepackage{color}
\usepackage{CJK}
\usepackage{multirow} 
\usepackage{subcaption}

\title{From Spelling to Grammar: A New Framework \\for Chinese Grammatical Error Correction}

\author{Xiuyu Wu$^1$$^,$$^2$ \ \ Yunfang Wu$^1$$^,$$^3$\thanks{*Corresponding author.}\\
   $^1$MOE Key Laboratory of Computational Linguistics, Peking University, Beijing, China\\
   $^2$School of Software and Microelectronics, Peking University, Beijing, China\\
   $^3$School of Computer Science, Peking University, Beijing, China\\
  {\tt \{xiuyu\_wu,wuyf\}@pku.edu.cn}
}

\begin{document}
\maketitle
\begin{abstract}
Chinese Grammatical Error Correction (CGEC) aims to generate a correct sentence from an erroneous sequence, where different kinds of errors are mixed. This paper divides the CGEC task into two steps, namely spelling error correction and grammatical error correction. Specially, we propose a novel zero-shot approach for spelling error correction, which is simple but effective, obtaining a high precision to avoid error accumulation of the pipeline structure. To handle grammatical error correction, we design part-of-speech (POS) features and semantic class features to enhance the neural network model, and propose an auxiliary task to predict the POS sequence of the target sentence. 
Our proposed framework achieves a 42.11 $F_0.5$ score on CGEC dataset without using any synthetic data or data augmentation methods, which outperforms the previous state-of-the-art by a wide margin of 1.30 points. Moreover, our model produces meaningful POS representations that capture different POS words and convey reasonable POS transition rules. 
\end{abstract}

\section{Introduction}
\label{sect:introduction}

Grammatical error correction (GEC) takes erroneous sequences as input and generates correct sentences. In recent years, English GEC task has attracted wide attention from researchers. By employing pre-trained models \cite{model_bertfuse,egec_bart_gatsumata_2020} or incorporating synthetic data \cite{syndata_Grundkiewicz_2019, syndata_lichtarge_2019}, 
the sequence-to-sequence models achieve remarkable performance on English GEC task. Besides, several sequence labeling approaches are proposed to cast text generation as token-level edit prediction~ 
\cite{LaserTagger_malmi_2019, PIE_Awasthi_2019, GECToR_omelianchuk_2020}.

Chinese grammatical error correction (CGEC) is less addressed. Previous works adopt ensemble methods by combining seq2seq networks with heuristic rules \cite{nlpcc_model_ali,nlpcc_model_youdao} or sequence editing approaches \cite{HRG_2020, mucgec_zhang_2022}. Different from English, 
Chinese language utilizes function words instead of affixes to represent forms and tenses, making it hard to design detailed Chinese-specific edit labels. The simple label strategy (for example, \emph{Keep}, \emph{Delete}, \emph{Append\_X}) has been proved not competitive with the seq2seq model on CGEC task \cite{span_detection_Chen_2020, mucgec_zhang_2022}.

\begin{figure}
    \centering
    \small
    \includegraphics[width=\linewidth]{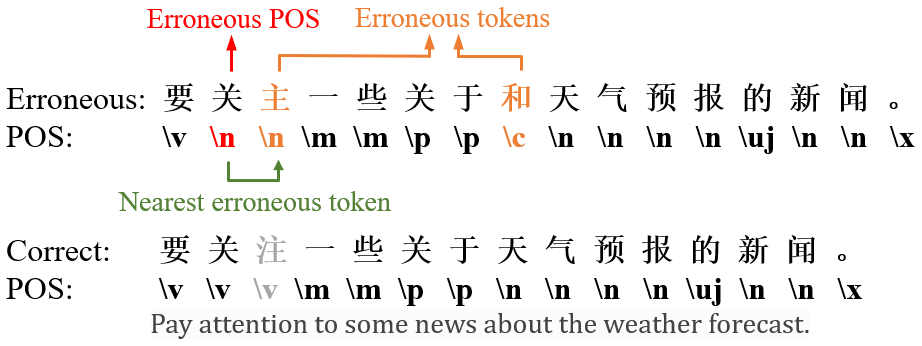}
    \caption{An example of erroneous-correct sentence pair. Black colour: correct tokens. Orange: erroneous tokens. Red: correct tokens with wrong POS tags.}
    \label{fig:feaute_analysis}
\end{figure}

Generally, the errors occurring in Chinese texts can be divided to spelling errors and grammar errors. For example in Figure \ref{fig:feaute_analysis}, "\begin{CJK}{UTF8}{gbsn}主\end{CJK}" is a spelling error which should be subtitued to "\begin{CJK}{UTF8}{gbsn}注\end{CJK}". While "\begin{CJK}{UTF8}{gbsn}和\end{CJK} (and)" relates to grammatical error which should be deteled. According to our statistics on HSK data\footnote{\url{http://hsk.blcu.edu.cn/Login}}, which is collected from the writing section of Chinese proficiency exam, the proportion of spelling errors is about 18.58\%. Unfortunately, most of previous works mix two kinds of errors and adopt the same model to handle them. 
Moreover, as the basic elements for sentence understanding and processing, spelling errors will influence the usage of high-level features in the CGEC task.    
Although amounts of linguistic features have been investigated to improve many natural language processing tasks, deep syntactic and semantic knowledge is rarely explored in CGEC.

Therefore, we propose a new framework for CGEC with two steps: 
\textbf{S}pelling error correction and semantic-enriched \textbf{G}rammatical error correction (SG-GEC). We propose a novel zero-shot method for Chinese spelling error correction, by taking advantage of the pre-trained BERT and Chinese phonetic information, which is straightforward but achieves a satisfying precision. Further, we introduce semantic knowledge into the seq2seq model to correct grammatical errors. We carefully analyze the reliability and utility of part-of-speech (POS) in erroneous-correct paired sentences, and design an effective method to integrate POS and semantic representations into the neural network model. Moreover,  
we introduce an auxiliary task of POS sequence prediction, where a Conditional Random Field (CRF) layer is added to ensure the valid of generated POS sequences and stimulate the model to learn grammar-level corrections.

We conduct extensive experiments on CGEC NLPCC dataset \cite{nlpcc_dataset}. Experimental results show that our proposed zero-shot spelling error correction module achieves a 60.25 precision, which lays a good foundation for further leveraging word-level features. With the pre-trained BART for initialization, our model achieves a new state-of-the-art result of 42.11 $F_{0.5}$ score, which outperforms all previous approaches including pre-trained models and ensemble methods. We also evaluate model performance on CGED-2020 test dataset \cite{ged_dataset_2020} and obtain satisfying results.


To sum, our contributions are as follows: 
\begin{itemize}
    \item We present a new framework for CGEC, which first conducts a preliminary spelling error correction and then performs grammatical error correction with semantic features. 
    \item We propose a novel zero-shot Chinese spelling error correction method, which is straightforward and 
    achieves a high precision.
    \item We effectively inject semantic knowledge to CGEC at both encoder and decoder, by incorporating POS and semantic class features into the input embeddings, and introducing an auxiliary task of POS sequence generation in the decoding phase. 
    \item Our proposed model obtains a new state-of-the-art result on CGEC task, outperforming previous works by a wide margin 
    without using any data augmentation method.
\end{itemize}

\section{Observation and Intuition}

\label{sect:feautre significance}
Various types of linguistic features have been exploited in NLP, which bring improvement on different tasks. However, it remains an open issue to 
introduce linguistic features to GEC. Different from other NLP tasks, the GEC task takes erroneous sentences as input,
based on which the extra features 
might bring noise 
to the GEC model that harms the performance.

\subsection{Part-of-Speech and Grammar Errors}
\label{sect:pos feature}
Part-of-speech represents the syntactic function of a word in contexts, which is closely connected with grammar. To bring POS features to the GEC task, the reliability and sensitivity of POS tags to grammar errors  
should be carefully examined. We conduct such analysis on NLPCC dataset, using Jieba\footnote{\url{https://github.com/fxsjy/jieba}} as the POS tagger. 

According to our statistics, 88.2\% of erroneous sentences have different POS sequences with their paired correct sentences, demonstrating that POS feature is sensitive to grammatical errors. 
An example is given in Figure \ref{fig:feaute_analysis}. 
We count LCS (Longest Common Sub-sequence) between erroneous-correct sentence pairs. We divide tokens in the erroneous sentence into two types: \emph{Corr-token} which appears in LCS and \emph{Err-token} which does not appear in LCS. 
Consequently, 98.1\% \emph{Corr-tokens} have correct POS tags, proving that the POS tagger could provide precise feature for the correct part of erroneous sentences. As for those 1.9\% \emph{Corr-tokens} which have wrong POS tags (red colour in Figure \ref{fig:feaute_analysis}), we calculate the average distance between them and the nearest \emph{Err-token} (orange color) and the result is 2.38 tokens. In contrast, the average distance between \emph{Corr-tokens} with correct POS tags and the nearest \emph{Err-token} is 8.59 tokens. This   
suggests that \emph{Corr-tokens} with wrong POS tags are next to the erroneous part of sentences. 

All these statistical results demonstrate that the POS feature is sensitive to the erroneous part meanwhile is robust for the correct part of sentences.

\begin{figure*}[t]
\centering
\includegraphics[width=14cm]{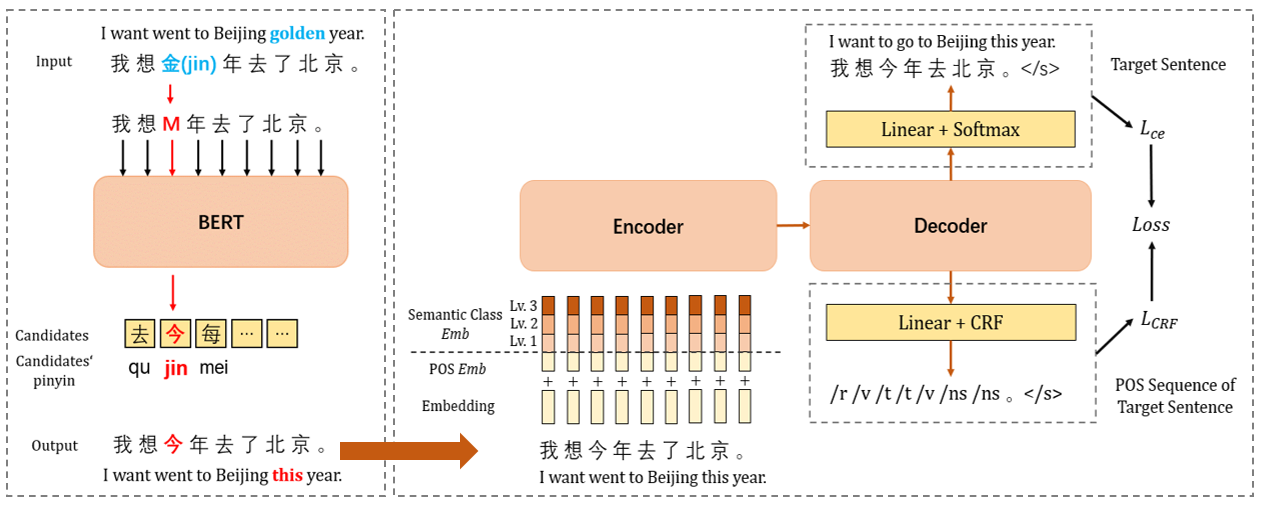}
\caption{Overview of our new framework for CGEC, which is composed of Spelling Error Correction (left part) and Grammatical Error Correction (right part). \textbf{M} refers to the [MASK] symbol in the BERT model.}
\label{overview of framework}
\end{figure*}

\subsection{Semantic Class and Grammar Errors}
Word semantic class is a kind of context-free feature, which tells which class each word belongs to according to a semantic dictionary. The dictionary is organized in a tree structure, consisting of different levels of semantic classes. For example, top-3 level semantic class of soup is entity, water and boiled water. By introducing semantic class knowledge, the model could learn the correlation between different semantic classes, and thus correct some 
semantic collocation errors, such as "\begin{CJK}{UTF8}{gbsn}冬阴功对外国人的喜爱\end{CJK} (Seafood soup enjoys foreigners.)" . In this example, a kind of food is incorrectly used as the subject which performs the action "enjoy" .

We leverage HIT-CIR Tongyici Cilin (Extended)~\footnote{\url{http://ir.hit.edu.cn/demo/ltp/Sharing_Plan.htm}} to provide semantic class knowledge.  

\section{Zero-shot Spelling Error Correction}
\label{sect:spell correct}

We present a new framework for CGEC as shown in Figure \ref{overview of framework}, which consists of Spelling Error Correction (SEC) and Grammatical Error Correction (GEC). Firstly, we propose a smart zero-shot method for SEC. 

Formally, the input erroneous sentence is represented as  $X = (x_1, x_2, ...x_n)$, and the target correct sentence is denoted as $Y = (y_1, y_2, ...y_m)$, where $n,m$ mean the length of sentence. We use $\tilde{X} = (\tilde{x}_1, \tilde{x}_2, ..., \tilde{x}_n)$ to represent the output of the zero-shot spelling error correction module:

\begin{equation}
    \tilde{X} = SEC(X)
\end{equation}

Specially, if a token ${x}_i$ in $X$ has a relatively high probability of being written incorrectly, it will be substituted with a [MASK] token. Then, the pre-trained BERT model \cite{model_bert} is employed to generate top-3 candidate tokens $V = (v_1, v_2, v_3)$ with high probability. Among the candidates, we select the token that most likely appears in the [MASK] position:


\begin{equation}
    \tilde{x}_i = \left\{
    \begin{aligned}
        x_i, 
        \quad & v_j \notin SimSet(x_i) \\
        v_j, \quad & v_j \in SimSet(x_i) \\
    \end{aligned}
    \right.
\end{equation}
If there exists more than one token belonging to $SimSet(x_i)$, we choose the token with the highest score generated by BERT. According to the previous study, over 80\% spelling errors in Chinese are related to phonological similarity \cite{simichar_2010}. So, we set $SimSet(x_i)$ to be the collection of homophones of $x_i$.

Figure \ref{overview of framework} gives an example in the left part. In the given sentence, \begin{CJK}{UTF8}{gbsn}金\end{CJK}(golden) is suspected to be written incorrectly and substituted with [MASK]. The top three tokens with the highest score generated by BERT are: 
\begin{CJK}{UTF8}{gbsn}去\end{CJK}(last), 
\begin{CJK}{UTF8}{gbsn}今\end{CJK}(this) and 
\begin{CJK}{UTF8}{gbsn}每\end{CJK}(every). Among them, 
\begin{CJK}{UTF8}{gbsn}今\end{CJK}/jin shares the same PINYIN with 
\begin{CJK}{UTF8}{gbsn}金\end{CJK}/jin. So we replace \begin{CJK}{UTF8}{gbsn}金\end{CJK} with \begin{CJK}{UTF8}{gbsn}今\end{CJK} in the original sentence.

A problem in SEC is how to decide whether a token $x_i$ is likely to be written incorrectly. Intuitively, the punctuation and commonly used Chinese characters, whose occurrences in the training dataset are over $k_c$, are less likely to be written incorrectly. We directly keep these tokens unchanged to improve the precision of SEC module and reduce the computational expense. 
To find out the appropriate threshold value $k_c$, we conduct experiments on the test set of SIGHAN-2015 \cite{sighan_dataset}, which is designed specially for Chinese spelling error correction and contains 1100 examples collected from Chinese language learners.

\begin{figure}[ht]
    \centering
    \includegraphics[width=0.7\linewidth]{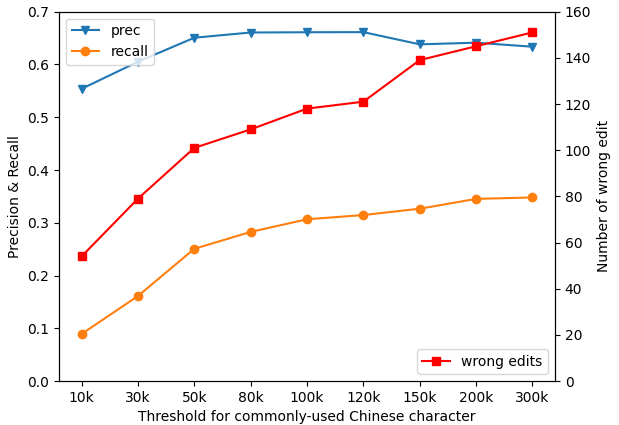}
    \caption{Result of zero-shot spelling correction module evaluated on SIGHAN test dataset.}
    \label{fig:sighan_sec}
\end{figure}

As shown in Figure \ref{fig:sighan_sec}, the precision score on SIGHAN test dataset has peaked at 66.1 when $k_c = 80,000$ and gradually declines after $k_c > 120,000$. 
In order to restrain error accumulation of the pipeline structure, we hope our SEC module to be high in precision. Accordingly, we set $k_c = 80,000$ in our experiment. 

\section{Integrating Semantics for Grammatical Error Correction}
\label{sect:model}
We adopt the Transformer encoder-decoder architecture for grammatical error correction. To project semantic knowledge to CGEC, at the encoder we incorporate the semantic knowledge representations, and at the decoding we design the POS sequence generation as an auxiliary task.  

We add the semantic knowledge embedding $E^{semantics}$ to the original word embedding to serve as the input of encoder: 

\begin{equation}
    H^{src} = Encoder(E^{word} + E^{semantics})
\end{equation}
In the decoding phase, we take the hidden state of timestep \emph{t} to predict the $t^{th}$ token in the target sentence:

\begin{equation}
    H^{tgt}_t = Decoder(H^{tgt}_{\leq t-1}, H^{src})
\end{equation}

\begin{equation}
    P^v_t(w) = softmax(Linear(H^{tgt}_t))
\end{equation}
where $P^v_t(w)$ is the generation probability of each token.

\subsection{Injecting Semantic Features}
\label{sect:feature embedding}
The semantic knowledge is composed of POS and semantic classes. Please note that, in this stage, the input sequence is $\tilde{X}$, where the spelling error correction has been conducted.

We leverages a POS tagger to obtain the POS tag $\tilde{X}^p = (\tilde{x}^p_1, \tilde{x}^p_2, ...\tilde{x}^p_n)$ for each token in $\tilde{X}$. The embedding of POS tag sequence is computed as:

\begin{equation}
    E^p_i = Emb^p(\tilde{x}^p_i), E^p_i \in \mathbb{R}^{d_p}
\end{equation}

As shown in Table 1, there are different levels of semantic classes to specify a word. 
We use $\tilde{X}^{c, l} = (\tilde{x}^{c,l}_1, \tilde{x}^{c,l}_2, ... \tilde{x}^{c,l}_n)$ to represent the $l^{th}$ level class feature for each token. The high level semantic class brings precise information. If only the high level class feature is extracted, the model will treat words as individual groups and ignore their relations in low levels. Therefore, the semantic representation of token $x_i$ is calculated as the concatenation of embeddings of the first \emph{k-th} levels:

\begin{equation}
    E^c_i = E^{c,1}_i \oplus E^{c,2}_i \dots \oplus E^{c,k}_i, E^c_i \in \mathbb{R}^{d_c \times k}
\end{equation}

\begin{equation}
    E^{c,l}_i = Emb^{c,1}(\tilde{x}^{c,l}_i), E^{c,l}_i \in \mathbb{R}^{d_c}
\end{equation}

Considering that POS could be regarded as a kind of rough semantic knowledge and be located at the lowest level of semantic class, we concatenate the POS embedding and semantic class embedding to obtain the semantic representation: 

\begin{equation}
    E^{semantics}_i = E^p_i \oplus E^c_i
\end{equation}

To make the dimension of semantic embedding equal to that of word embedding $d_E$, the dimension of POS embedding $d_p$ and that of semantic class embedding $d_c$ are set to:

\begin{equation}
    d_p = d_c = \frac{d_E}{k + 1}
\end{equation}

\subsection{Predicting POS Sequence}
\label{sect:pos correction}
As described in Section \ref{sect:pos feature}, the wrong POS tags are usually close to the erroneous parts of sentences, which indicates that token-level error correction shares the same target with POS-level error correction. Moreover, POS-level errors are more general, 
since various types of token-level errors might be mapped to the same on POS-level. Inspired by this observation, we design a sub-task to predict the error-free POS sequence.

At timestep \emph{t}, the generation probability of each token's POS tag is computed utilizing the linear function and softmax:

\begin{equation}
P^{pos}_t(w) = softmax(Linear(H^{tgt}_t)) 
\end{equation}

The cross entropy loss is commonly used to stimulate the model to generate a target sequence. However, besides being close to the golden correct POS sequence, the generated POS sequence itself should be well-formed. 
To model the dependencies among neighboring POS tags, we adopt Conditional Random Fields (CRF) \cite{crf_lafferty_2001}, under which the likelihood of target POS sequence $Y^p = (y^p_1, y^p_2, ..., y^p_m)$ is computed as:

\begin{equation}
    \begin{aligned}
    P_{crf}^{pos}(Y^p|X) & = \\ 
    \frac{1}{Z(X)} & exp\Big( \sum_{t=1}^ms(y^p_t) + \sum_{t=2}^mt(y^p_{t-1}, y^p_{t}) \Big)
    \end{aligned}
\end{equation}
where $s(y^p_t) = P^{pos}_t(w)$, which represents the generation probability of $y^p_t$.

The value $t(y^p_{t-1}, y^p_t) = M_{y^p_{t-1},y^p_t}$ denotes the transition score from POS tag $y^p_{t-1}$ to $y^p_t$, which can be learnt as parameters during the end-to-end training procedure. The Viterbi algorithm \cite{viterbi_Forney_1973, crf_lafferty_2001} is utilized to calculate the normalizing factor $Z(X)$.

\subsection{Training Objective}
As shown in Figure \ref{overview of framework},
our model is trained to generate the target sentence and POS sequence simultaneously, and thus the final loss is computed as:

\begin{equation}
    Loss = L_{ce} + L_{CRF} \\
\end{equation}

\begin{equation}
    L_{ce} = -log\sum_{j=1}^mP^{v}(y_j)
\end{equation}

\begin{equation}
    L_{CRF} = - logP_{crf}^{pos}(Y^p|X)
\end{equation}


\section{Experimental Setup}
\label{sect:experiment}

\subsection{Dataset and Evaluation Metric}
We conduct experiments on the dataset of NLPCC-2018 shared task \cite{nlpcc_dataset}
which contains 1.12 million training samples collected from the language learning platform Lang-8 \footnote{\url{https://lang-8.com/}} and 2000 human annotated samples for test.
We randomly selected 5,000 instances from training data as the development set. Besides, we changed the format of CGED-2020 test dataset \cite{ged_dataset_2020} to suit our task, and manually corrected 283 word-order errors in CGED-2020 to obtain error-free sentences (Please refer to Appendix A). We evaluate our model on CGED-2020 (1457 samples) as a supplement. 

For NLPCC-2018 test dataset, we segment model outputs by the official PKUNLP tool, and adopt the official MaxMatch ($M^2$) \cite{m2score} scorer
to calculate precision, recall and $F_{0.5}$ score.
For CGED-2020 test dataset, we apply the simple char-based evaluation using ChERRANT~\footnote{\url{https://github.com/HillZhang1999/MuCGEC/tree/main/scorers/ChERRANT}} to avoid the influence brought by different word segmentation tools.

\subsection{Training Details}
Our model is implemented using Fairseq.
We average parameters of the last 5 checkpoints. We use BART-base-chinese\footnote{\url{https://huggingface.co/fnlp/bart-base-chinese}} to initialize our model. We use BERT tokenizer for word tokenization and replace some [unused] tokens with Chinese punctuation. 
Please refer to Appendix \ref{app:hyperparamters} for more parameter settings.

\begin{table*}[t]
\centering
\small
\begin{tabular}{l|ccc|ccc}
    \toprule
    \multirow{2}{*}{Model} &
    \multicolumn{3}{c|}{NLPCC-2018} &
    \multicolumn{3}{c}{CGED-2020} \\
    \multicolumn{1}{l|}{} & P & R & $F_{0.5}$ & P & R & $F_{0.5}$ \\
    \midrule
    \midrule
    AliGM$\blacktriangle$ \cite{nlpcc_model_ali} & 41.00 & 13.75 & 29.36 & - & - & - \\ 
    YouDao$\blacktriangle$ \cite{nlpcc_model_youdao} & 35.24 & 18.64 & 29.91 & - & - & - \\
    BLCU$\blacktriangle$ \cite{nlpcc_model_blcu}  & 47.63 & 12.56 & 30.57 & - & - & - \\ 
    ESD-ESC \cite{span_detection_Chen_2020} & 37.30 & 14.50 & 28.40 & - & - & - \\ 
    S2A model \cite{s2a_model_2022} & 36.57 & 18.25 & 30.46 & - & - & - \\
    HRG$\blacktriangle$ \cite{HRG_2020} & 36.79 & \textbf{27.82} & 34.56 & - & - & - \\
    MaskGEC \cite{maskgec_wang} & 44.36 & 22.18 & 36.97 & - & - & - \\
    \midrule
    Transformer & 38.43 & 12.95 & 27.58 & 31.69 & 11.43 & 23.39 \\
    SG-GEC (Transformer) & 44.52 & 18.28 & 34.59 & 32.37 & 12.04 & 24.20 \\
    \midrule
    BERT-fuse \cite{model_bertfuse} & 42.01 & 20.24 & 34.57 & 31.50 & 14.99 & 25.81 \\
    GECToR~\cite{mucgec_zhang_2022} & 39.83 & 23.01 & 34.75 & - & - & - \\
    GECToR (Our implement) & 38.76 & 23.19 & 34.17 & 33.33 & 19.46 & 29.17 \\
    3$\times$Seq2Edit + 3$\times$Seq2Seq$\blacktriangle$ \cite{mucgec_zhang_2022} & \textbf{55.58} & 19.78 & 40.81 & - & - & - \\
    BART & 46.21 & 25.14 & 39.58 & 38.89 & \textbf{20.13} & 32.78  \\
    BART + MaskGEC & 48.79 & 24.03 & 40.45 & 40.72 & 18.63 & 32.91 \\
    SG-GEC (BART init) & 50.56 & 25.24 & \textbf{42.11} & \textbf{40.97} & 20.05 & \textbf{33.90} \\
    \bottomrule
\end{tabular}
\caption{\label{tab:overall performance} Performance comparison on the NLPCC-2018 test dataset~\cite{nlpcc_dataset} and CGED-2020 test dataset~\cite{ged_dataset_2020}. $\blacktriangle$ refers to ensemble model.}
\end{table*}

\subsection{Comparing Methods}
We compare our model with \textbf{YouDao} \cite{nlpcc_model_youdao}, \textbf{AliGM} \cite{nlpcc_model_ali} and \textbf{BLCU} \cite{nlpcc_model_blcu} , which are the three top systems in the NLPCC-2018 challenge.

Also, the following previous works are referred as baseline models:

\textbf{ESD-ESC} uses a pipeline structure to firstly detect the erroneous spans and then generate the correct text for annotated spans \cite{span_detection_Chen_2020}.

\textbf{HRG} proposes a heterogeneous approach composed of a LM-based spelling checker, a NMT-base model and a sequence editing model \cite{HRG_2020}.

\textbf{MaskGEC} adds random noises to source sentences dynamically in the training process \cite{maskgec_wang}. 

\textbf{S2A model} combines the output of seq2seq framework and token-level action sequence prediction module \cite{s2a_model_2022}.

More recently, \cite{mucgec_zhang_2022} enhance the text editing model \textbf{GECToR} \cite{GECToR_omelianchuk_2020} by using Struct-BERT as its encoder. They also ensemble GECToR with fine-tuned BART model (denotes as \textbf{3$\times$Seq2Edit + 3$\times$Seq2Seq}) utilizing edit-wise vote mechanism. 

Besides, we finetune \textbf{BART} \cite{model_bart} on training dataset and apply MaskGEC as data augmentation method to provide a strong baseline.


\section{Results and Analysis}
\label{sect:result}

\subsection{Overall Performance}
Table \ref{tab:overall performance} reports the main evaluation results of our proposed model on NLPCC-2018 and CGED test datasets, comparing with previous researches. 

Our proposed SG-GEC model obtains a new state-of-the-art result with a 42.11 $F_{0.5}$ score, which outperforms the previous best single / ensemble model by 5.14 / 1.30 points. Meanwhile, our SG-GEC model surpasses GECToR, which achieves SOTA result on English GEC task. Comparing with the base BART fine-tuned method, our strategy brings a performance gain of 2.53 points. What's more, our SG-GEC model achieves a significant better result in $precision$ among singe models, which is vital for some real-world applications. Without using pre-trained language models, our method outperforms the baseline Transformer by a large margin of 7.01 $F_{0.5}$ points.

Meanwhile, when being initialized by the pretrained BART, our proposed framework obviously surpasses MaskGEC. It demonstrates that our SG-GEC model brings additional semantic knowledge which is more beneficial to the strong BART model than simple data augmentation methods.

Our model consistently outperforms all other models when evaluated on CGED-2020 test dataset, which proves the generality of our model.

\begin{table}
\centering
\small
\begin{tabular}{lcccc}
    \toprule[1pt]
    Model & P & R & $F_{0.5}$ & Imp. \\
    \midrule
    \midrule
    SG-GEC (BART init) & 50.56 & 25.24 & 42.11 & - \\
    - SEC & 49.70 & 22.30 & 39.90 & - 2.21  \\
    - POS \emph{emb} & 48.85 & 25.95 & 41.52 & - 0.59 \\
    - Semantic Class \emph{emb} & 48.73 & 25.92 & 41.44 & - 0.67 \\
    - POS predict \& CRF  & 49.50 & 25.02 & 41.40 & - 0.71 \\
    - CRF & 50.03 & 25.21 & 41.80 & - 0.31 \\
    \bottomrule[1pt]
\end{tabular}
\caption{\label{tab:ablation study} Ablation study of our model on NLPCC dataset. \textbf{- CRF} refers to substitute crf loss with cross entropy loss.} 
\end{table}

\subsection{Ablation Study}
We conduct ablation study on NLPCC dataset to evaluate the effect of each module, as shown in Table \ref{tab:ablation study}. All the modules bring improvement in model performance. Specially, removing spelling error correction (SEC) results in a sharp decrease of 2.21 $F_{0.5}$ score, for the reason that it not only corrects spelling errors but also offers more reliable POS and semantic class features. Setting embedding of POS / semantic class features to zero leads a decrease of 0.59 / 0.67 $F_{0.5}$ score, which demonstrates that POS and semantic class features bring valid information for grammar error detection. Replacing CRF loss with cross entropy loss leads to a decrease of 0.31 point. Without the sub-task of POS generation, the model performance drops from 42.11 to 41.40. It illustrates that predicting POS with CRF layer helps the model to learn grammar-level correction which further improves the performance. 

\begin{table}
\centering
\small
\begin{tabular}{lcccc}
    \toprule[1pt]
    Model & \emph{Num.} & P & R & $F_{0.5}$\\
    \midrule
    \midrule
    SEC & 192 & 60.25 & 5.18 & 19.27 \\
    \midrule
    BART & 120 & 46.21 & 25.14 & 39.58 \\
    BART + SEC & 210 & 47.70 & 25.80 & 40.78 \\
    \midrule
    BART + SemF & 113 & 48.11 & 23.24 & 39.63 \\
    BART + SemF + SEC & 210 & 49.50 & 25.02 & 41.40 \\
    \midrule
    \midrule
    B-sec & 153 & 25.63 & 7.63 & 17.41 \\
    BART + B-sec & 174 & 38.98 & 25.48 & 35.21 \\
    BART + SemF + B-sec & 176 & 39.07 & 24.33 & 34.85 \\
    \bottomrule[1pt]
\end{tabular}
\caption{\label{tab:sec effect} Effect of spelling error correction. \emph{Num.} refers to the number of corrected spelling error tokens. 
\textbf{SEC} refers to zero-shot spelling error correction module.
\textbf{B-sec} is a BERT model finetuned on the spelling error correction dataset SIGHAN15 and HybirdSet.
\text{+ SemF} denotes integrating semantic features.} 
\end{table}

\subsection{Effect of Spelling Error Correction}
In our framework, zero-shot spelling error correction (SEC) is a vital step. We conduct further experiments to illustrate the effect of this module, and list the results in Table \ref{tab:sec effect}. 

Our proposed SEC module greatly improves the number of corrected spelling errors, with 90 more tokens over BART and 97 more tokens over the semantic-enriched BART. During the pre-training process of BART model, input tokens are substituted to [MASK] symbols and new tokens are generated without special constraints. Meanwhile, our SEC module intentionally masks misspelled tokens and takes phonetic similarity as constraints when generating new tokens, therefore corrects more spelling errors and achieves high precision score.



If semantic feature embeddings are directly added on BART 
without utilizing the SEC module, the number of corrected spelling errors will drop from 120 to 113. Because spelling errors influence word segmentation and thus lead to erroneous POS and semantic class features at the position of misspelled tokens. 
In contrast, our high precision SEC lays a solid foundation for the further semantic information injection. After applying the SEC module, BART + SEC + SemF (semantic features) obtains larger improvement in model performance.

We also compare our zero-shot SEC module with a BERT model finetuned on the spelling error correction dataset SIGHAN-2015 \cite{sighan_dataset}. Our SEC module strictly focuses on correcting spelling errors and achieves a high precision. However, beside spelling errors, the finetuned BERT model automatically corrects other types of errors, leading to a high recall but low precision score. As the first step of pipeline structure, the low precision brings huge noise to the subsequent module and thus damages the final performance.

\subsection{Analysis on POS representations}
The part-of-speech feature is closely connected with grammar. In our model, we inject POS embedding in encoder and predict the correct POS sequence in decoder, which enable our model to learn a better POS representation.

To investigate the POS representations, we calculate the nearest neighbours to each POS tag by computing the cosine distance between embedding vectors, and list the results in Table \ref{tab:similar tokens}. 
For each POS tag, most of their nearest tokens have the corresponding part-of-speech. It demonstrates that our POS embedding could capture general features of tokens sharing the same part-of-speech, which benefits our model and shows potential for other NLP applications. 

\begin{table}[h]
\centering
\small
\begin{CJK}{UTF8}{gbsn}
\begin{tabular}{l|lll}
    \toprule[1pt]
    POS Tag & \multicolumn{3}{c}{Top-3 nearest tokens} \\
    \midrule
    \midrule
    noun & 栈 storehouse & 障 obstacle & 浆 liquid\\
    verb & 想 think & 离 leave & 做 do  \\
    adjective & 脆 crisp & 傻 foolish & 幸 lucky \\
    adverb & 再 again & 也 also & 永 always \\
    pronoun & 我 I & 他 he & 飞 fly \\
    preposition & 对 for/towards & 把 \emph{prep.}  & 为 for \\
    \bottomrule[1pt]
\end{tabular}
\end{CJK}
\caption{\label{tab:similar tokens}Top-3 nearest tokens to POS tags.}
\end{table}

\begin{figure}[h]
    \centering
    \includegraphics[width=0.8\linewidth]{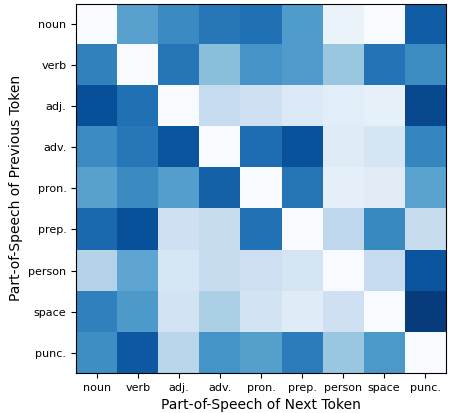}
    \caption{POS transition matrix in the CRF layer. Darker colour refers to higher transition probability.}
    \label{fig:crf visualize}
\end{figure}

In our model, CRF is essential to capture neighboring POS dependencies of target sequences. We visualize the POS transition matrix that the CRF layer has learnt in Figure \ref{fig:crf visualize}. Interestingly, several grammar rules could be found. For example, $preposition$ is usually followed by $noun$, $pronoun$ or $space name$, but it has a low probability of transiting to punctuation (the end of a sentence).
$Adjective$ usually occurs before $noun$ but seldom connects with $preposition$. 
This POS knowledge enables our model to generate grammatical sentences.

\subsection{Case Study}

Table \ref{tab:case} provides an example output of our SG-GEC model comparing with BART. Our SEC module firstly corrects the spelling error in this sentence. Benefiting by this, seq2seq model corrects the grammatical error subsequently. However, the BART model fails to correct both spelling and grammatical errors in this sentences. More cases are listed in Appendix \ref{app:more_case}.

We list two cases in Table \ref{tab:semantic_case} to show the effect of our semantic class feature. When object and verb are mismatched (Case 1) or verb is missing (Csed 2), our model could correct these errors benefited from information provided by semantic class feature. When meeting rarely used words, for example \begin{CJK}{UTF8}{gbsn}罪魁祸首\end{CJK}(chief culprit), semantic class feature might provide extra information learning from examples which contain \begin{CJK}{UTF8}{gbsn}主犯\end{CJK}(principal criminal) or \begin{CJK}{UTF8}{gbsn}要犯\end{CJK}(important crimial) and help model to replace verb \begin{CJK}{UTF8}{gbsn}
了解\end{CJK}(know about) with \begin{CJK}{UTF8}{gbsn}了结\end{CJK}(kill).

\begin{CJK}{UTF8}{gbsn}
\begin{table}[t]
    \small
    \centering
    \begin{tabular}{lp{5.5cm}}
        \toprule[1pt]
         Type & Sample  \\
         \midrule
         \midrule
         SRC & 由于羊驼毛的价格比羊羔毛低廉且更具优的保暖性...\\
         TGT & 由于羊驼毛的价格比羊羔毛低廉且更具\textcolor{red}{有}\textcolor{blue}{(delete-的)}保暖性... \\
         BART & 由于羊驼毛的价格比羊羔毛低廉且更具优的保暖性... \\
         SEC & 由于羊驼毛的价格比羊羔毛低廉且更具\textcolor{red}{有}的保暖性... \\
         SG-GEC &  由于羊驼毛的价格比羊羔毛低廉且更具\textcolor{red}{有}\textcolor{blue}{(delete-的)}保暖性... \\
         Translation & Because paco wool is cheaper and warmer than lamb wool... \\
         \bottomrule[1pt]
    \end{tabular}
    \caption{Case study of our model. Red / blue color refers to correction of spelling / grammatical error.}
    \label{tab:case}
\end{table}
\end{CJK}

\begin{CJK}{UTF8}{gbsn}
\begin{table*}[t]
    \centering
    \small
    \begin{tabular}{lp{13cm}}
        \toprule[1pt]
         Type & Sample  \\
         \midrule
         \midrule
         keyword & 罪魁祸首-chief culprit \\
         group words & 主犯-principal criminal 要犯-important criminal 正凶-principal murderer \\
         SRC & 我就在此\textcolor{blue}{了解(know about)}了你这\textcolor{red}{罪魁祸手(chief culprit)}，平平风气！ \\
         TGT & 我就在此\textcolor{green}{了结(kill)}了你这\textcolor{red}{罪魁祸首(chief culprit)}，平平风气！ \\
         BART & 我就在此\textcolor{blue}{了解(know about)}了你这\textcolor{red}{罪魁祸手(chief culprit)}，平平风气！ \\
         SG-GEC & 我就在此\textcolor{green}{了结(kill)}了你这\textcolor{red}{罪魁祸首(chief culprit)}，平平风气！\\
         Translation & I'm going to kill you chief culprit right here to improve social climate. \\
         \midrule
         keyword & 年级-grade \\
         group words & 班级-class 高年级-senior year \\
         SRC & 时间过的很快，我已经\textcolor{blue}{变成(become)}了\textcolor{red}{三年级(the third grade)}。 \\
         TGT & 时间过得很快，我已经\textcolor{blue}{变成(become)}了\textcolor{green}{三年级的学生(third-grade student)}。 \\
         BART & 时间过的很快，我已经\textcolor{blue}{变成(become)}了\textcolor{red}{三年级(the third grade)}。 \\
         SG-GEC &  时间过的很快，我已经\textcolor{green}{上(is in)}\textcolor{red}{三年级(the third grade)}了。 \\
         Translation & Time passed quickly, I have become a third-grade student. \\
         \midrule
         keyword & 意思-meaning \\
         group words & 意义-significance 含义-implication \\
         SRC &  背完生词之后，再读课文，那么更容易生词的用法和\textcolor{red}{意思(meaning)}。 \\
         TGT & 背完生词之后，再读课文，那么更容易\textcolor{green}{记住(remember)}生词的用法和\textcolor{red}{意思(meaning)}。 \\
         BART &  背完生词之后，再读课文，那么生词的用法和\textcolor{red}{意思(meaning)}就更容易了。 \\
         SG-GEC &  背完生词之后，再读课文，会更容易\textcolor{green}{理解(understand)}生词的用法和\textcolor{red}{意思(meaning)}。 \\
         Translation & Reading the passage after reciting the new words makes it easier to remember the usage and the meaning of the new words。 \\
         \bottomrule[1pt]
    \end{tabular}
    \caption{Case study of our model. Group words refer to words which share the same semantic class with the keyword. Blue / red color refers to verb / object. Green color refers to modification of verb or object.}
    \label{tab:semantic_case}
\end{table*}
\end{CJK}

\section{Related Work}
\subsection{Grammatical Error Correction}
Seq2seq generation model and edit label prediction model are two mainstream models for GEC task. Benefiting by the rapid gains in hardware and high quality dataset, Transformer-based seq2seq models \cite{gec_transformer_Junczys_2018, egec_bart_gatsumata_2020, model_bertfuse} outperform traditional CNN and RNN-based model structures \cite{rnn_Xie_2016, rnn_Yuan_2016, cnn_Chollampatt_2018}. Copy mechanism and subtask is also introduced to seq2seq model \cite{copy_augmented_zhao}. LaserTagger \cite{LaserTagger_malmi_2019} treats the GEC task as text edit task and predicts \emph{Keep}, \emph{Delete} and \emph{Append\_\#} for each token in erroneous sentences to represent different edit operation. PIE \cite{PIE_Awasthi_2019} and GECToR \cite{GECToR_omelianchuk_2020}
manually design detailed English-specific labels, regarding case and tense. Synthetic data is generated to enhance model performance \cite{syndata_ge_2018, syndata_Grundkiewicz_2019, syndata_lichtarge_2019}. Besides two mainstream model structure, ESD-ESC \cite{span_detection_Chen_2020} firstly detects erroneous spans and generates correct contents only for annotated spans. TtT model \cite{tail2tail_2021} directly predicts each tokens in correct sentences given erroneous sentence.

CGEC task is less addressed. Release of NLPCC-2018 dataset \cite{nlpcc_dataset} attracts much attention from participated teams, where top 3 systems are AliGM \cite{nlpcc_model_ali}, YouDao \cite{nlpcc_model_youdao} and BLCU \cite{nlpcc_model_blcu}. HRG combines spelling checker, NMT-base model and sequence editing model \cite{HRG_2020}. However, spelling checker in HRG is based on language model which could not make full use of context. \citeauthor{maskgec_wang} proposed data augmentation method MaskGEC, which adds random noise to input sentence dynamically in training process. S2A model combines seq2seq and sequence editing model by combining 
prediction probability of words and edit labels \cite{s2a_model_2022}. \citeauthor{mucgec_zhang_2022} ensembles seq2seq model and sequence editing model by edit-wise vote mechanism and achieves the state-of-the-art on NLPCC-2018 dataset.

\subsection{Chinese Spelling Error Correction}
Chinese spelling error correction is firstly tackled with CRF or HMM models \cite{NTOU_tseng_2015, hanspeller_zhang_2015}. 
In recent neural network models, phonological and graphic knowledge is introduced to help detecting and correcting Chinese spelling errors \cite{sec_faspell_2019, sec_phmospell_2021, spellgcn_cheng_2020}. The pre-trained BERT model is also utilized to generate candidate sentences \cite{sec_faspell_2019, sec_softbert_zhang}.


Different from these models, we locate possibly misspelled tokens based on rule instead of neural network. We directly choose the homophone of masked token from candidates generated by BERT. Knowledge is utilized more explicitly in our module. More importantly, our method is zero-shot, without using any labeled data.

\section{Conclusion}
In this paper, we divide CGEC into two consecutive tasks: spelling error correction and grammatical error correction. We propose a zero-shot spelling error correction method, utilizing the pre-trained BERT model and taking advantage of Chinese phonological knowledge. It achieves a high precision score to avoid error accumulation in the pipeline structure.
Based on the careful analysis on real data, we inject proper semantic features into the encoder. And at the same time, we generate correct POS sequence as a sub-task to help generate correct sentences, where CRF is applied to guarantee the validness of the generated POS sequence.
Initialized by the pre-trained BART model, our proposed framework achieves a new state-of-the-art result on CGEC task, outperforming the previous best result by a large margin. 

\section*{Limitation}
Our zero-shot spelling error correction module is specifically designed for Chinese language. Meanwhile, the POS tagger and vocabulary of semantic class we used in SG-GEC model cannot be directly applied to other languages. To some degree, it makes SG-GEC model as a language-specific model. We try to find matched resources in English language and conduct experiments on English GEC dataset. The result is reported in Appendix \ref{app:eng_experiment}. It demonstrates that introducing semantic features after spelling check and employing sub-task of POS correction with CRF layer, which is the main idea of our work, could benefit GEC task of other languages.

\section*{Acknowledgement}
This  work  is  supported  by  the  National  Natural  Science  Foundation  of  China  (62076008),  the  Key Project of Natural Science Foundation of China (61936012) and the National Hi-Tech RD Program of China (No.2020AAA0106600).

\bibliography{Wu_EMNLP_2022}
\bibliographystyle{acl_natbib}

\appendix

\section{Annotation of CGED-2020 Dataset}
\label{app:word-order corr}
There exist 283 word-order errors in CGED-2020 test dataset. We first correct other three types of errors in the sentences according to the golden answer and mark the start and end points of word-order errors. Two annotators are asked to correct the error by adjusting the order of the tokens between start and end point. For 87\% of erroneous sentences, correction results of two annotators are consistent with each other. For the other 13\% sentences, a third annotator is asked to select the better one from two different corrected sentences. 

\section{Hyperparameters}
\label{app:hyperparamters}
The detailed hyperparameter settings are listed in Table \ref{tab:hyperparameters}.

\begin{table}[th]
\centering
\small
\begin{tabular}{lc}
    \toprule[1pt]
    Hyper-parameters & Value \\
    \midrule
    pretrained model & bart-base-chinese \\
    dropout & 0.1 \\
    learning rate & 3e-5 \\
    optimizer & Adam($\beta_1$=0.9, $\beta_2$=0.999, $\epsilon$=1e-8) \\
    lr scheduler & polynomial decay \\
    warmup updates & 500 \\
    total number updates & 20000 \\
    max tokens & 4096 \\
    update-freq & 2 \\
    max epochs & 10 \\
    loss function & cross entropy \\
    beam size & 12 \\
    \midrule
    Num. of parameters & 117 millions \\
    device & two NVIDIA RTX 2080 GPUs \\
    runtime & 4.5 hours \\
    \bottomrule[1pt]
\end{tabular}
\caption{\label{tab:hyperparameters} Hyper-parameter settings in our model.}
\end{table}

\section{Effect of Semantic Features}

We investigate the effect of each single feature as well as different approaches for feature combination, and the results are shown in Table \ref{tab:feature effect}. Both POS and semantic class bring helpful knowledge for GEC task. The Level-3 semantic class feature outperforms other single features, which provides more detailed classification information of semantic knowledge. Compared with accumulating, concatenating all features obtains a slightly better result.

\begin{table}[h]
\centering
\small
\begin{tabular}{lccccc}
    \toprule[1pt]
    Model & Num. & P & R & $F_{0.5}$ & Imp. \\
    \midrule
    \midrule
    BART + SEC & - & 47.70 & 25.80 & 40.78 & \\
    + \emph{POS} & 44 & 49.17 & 25.07 & 41.24 & + 0.46 \\
    + \emph{Class Lv.1} & 18 & 48.78 & 25.44 & 41.22 & + 0.44 \\
    + \emph{Class Lv.2} & 101 & 48.91 & 25.50 & 41.32 & + 0.54 \\
    + \emph{Class Lv.3} & 1431 & 49.50 & 24.91 & 41.34 & + 0.56\\ 
    \midrule
    + \emph{accum. All} & - & 49.01 & 25.50 & 41.38 & + 0.60 \\
    + \emph{concat. All} & - & 49.50 & 25.02 & 41.40 & + 0.62 \\ 
    \bottomrule[1pt]
\end{tabular}
\caption{\label{tab:feature effect} Effect of different semantic features. 
\emph{Class Lv.k} refers to the \emph{k-th} level semantic class. \emph{accum. All} / \emph{concat. All} denotes the final semantic representation is obtained by accumulating / concatenating all features' embeddings. \textbf{Num.} refers to number of different values of a specific feature.}
\end{table}

\section{Effect of Semantic Sequence Prediction}

\begin{table}[h]
\centering
\small
\begin{tabular}{lccccc}
    \toprule[1pt]
    Model & P & R & $F_{0.5}$ & Imp. \\
    \midrule
    \midrule
    
    BART+SEC+SemF & 49.50 & 25.02 & 41.40 & \\
    + \emph{POS pred} &  50.56 & 25.24 & 42.11 & + 0.71 \\
    \quad \emph{w/o CRF} & 50.03 & 25.21 & 41.80 & + 0.40 \\
    + \emph{Class Lv.1 pred} & 49.57 & 24.70 & 41.41 & + 0.01 \\
    \quad \emph{w/o CRF} &  49.57 & 24.70 & 41.26 & - 0.14 \\
    + \emph{Class Lv.2 pred} & 49.20 & 24.85 & 41.19  & - 0.21 \\
    \quad \emph{w/o CRF} & 50.29 & 23.11 & 40.71 &  - 0.69 \\
    \bottomrule[1pt]
\end{tabular}
\caption{\label{tab:prediction effect} Effect of different types of sequence generation as a sub-task. \emph{POS pred} / \emph{Class Lv.k pred} refers to employ a sub-task to predict POS / \emph{k-th} level semantic class sequence. \emph{w/o CRF} represents the standard cross entropy loss is applied without using CRF.}
\end{table}

To evaluate the effect of our proposed multitask learning framework, we exploit different semantic sequences as targets of generation, and the results are shown in Table \ref{tab:prediction effect}. Employing POS sequence generation as a sub-task outperforms other auxiliary tasks of semantic class sequence generation, which brings a further performance gain over the strong model of BART+SEC+SemF. POS sequence is more generalized and convey much more syntactic information. By learning to generate correct POS sequence, model could learn grammar-level correction instead of token-level correction.
Predicting semantic class sequence does no benefit the performance of model. Firstly, semantic class is a context-free feature based on words, which means predicting sequence semantic class roughly equals to predicting sequence of token. What's more, there are about 15\% words in target sentences having no semantic class in the lexicon, which will influence the training process. 


\section{Experiment on English GEC dataset}
\label{app:eng_experiment}

For English GEC task, following \citeauthor{bea_2019_task} \shortcite{bea_2019_task}, we use Lang-8 Corpus of Learner English \cite{lang8_dataset}, FCE \cite{fce_dataset}, NUCLE \cite{nucle_dataset} and W\&I+LOCNESS \cite{bea_2019_task} as training data, CoNLL-2013 test set as dev set and evaluate on CoNLL-2014 \cite{conll14_dataset} test set.

In Chinese language, token might be incorrectly written as its homophone. Meanwhile, in English language, spelling mistakes usually caused by missing or mis-writing letters. Our phonological knowledge based zero shot-spelling error correction module could not be directly applied to English language. Spelling errors in English language cause out-of-vocabulary words, which makes it easier to be detected and corrected compared with Chinese. Therefore, we simply utilize a spelling checker\footnote{\url{https://github.com/barrust/pyspellchecker}} based on dictionary and edit distance to substitue zero-shot SEC module in SG-GEC.
 
We use NLTK\footnote{\url{https://www.nltk.org/}} as POS tagger. For semantic class knowledge, we could not find exactly matched resources in English language. We design two alternative solutions:
\begin{itemize}
    \item \textbf{zero-class-feature} We set the embedding of semantic class features to zero during both training and inference process.
    \item \textbf{Wordnet-class-feature} We use WordNet\footnote{\url{https://wordnet.princeton.edu/}} to get the semantic class features of a word by recursively searching the hypernym of this word. Number of values in \emph{1rt} / \emph{2nd} / \emph{3rd} level semantic class is 148 / 685 / 9852.
\end{itemize}

We use BART-base\footnote{\url{https://huggingface.co/facebook/bart-base}} to initialize our model.

\begin{table}[ht]
\centering
\small
\begin{tabular}{lccccc}
    \toprule[1pt]
    Model & P & R & $F_{0.5}$ \\
    \midrule
    \midrule
    spelling checker & 56.07 & 2.94 & 12.14 \\
    \midrule
    BART & 70.10 & 40.16 & 61.00 \\
    BART + spelling checker & 69.72 & 41.27 & 61.27 \\
    \midrule
    SG-GEC (BART init)  \\
    \quad zero-class-feature & 68.80 & 43.96 & 61.82 \\ 
    \quad Wordnet-class-feature & 69.01 & 44.07 & 62.00 \\
    \bottomrule[1pt]
\end{tabular}
\caption{Experiment on English GEC dataset.} 
\label{tab:eng_experiment} 
\end{table}

Table \ref{tab:eng_experiment} demonstrates that spell checker brings little benefit to BART-finetuned model on English GEC task. One reason is that spelling error in English causes out-of-vocabulary words, which is easily to detect. As shown in Table \ref{tab:eng_spelling_errors}, misspelled out-of-vocabulary words are usually divided into BPE level in BART model.

\begin{table}[ht]
\centering
\small
\begin{tabular}{lll}
    \toprule[1pt]
    Correct & Misspelled & Tokenized \\
    \midrule
    \midrule
    potential & potetial & pot \#\#et \#\#ial \\
    responsibility & resposiblity & resp \#\#os \#\#ibl \#\#ity \\
    hundreds & hundrends & h \#\#und \#\#rend \#\#s \\
    \bottomrule[1pt]
\end{tabular}
\caption{\label{tab:eng_spelling_errors} Examples of misspelled words in English GEC dataset.} 
\end{table}

By introducing POS feature and sub-task of POS correction with CRF layer while setting embedding of semantic class features to zero, our model achieves 61.82 $F_{0.5}$ score, which outperforms BART model. Semantic class features provided by Wordnet also slightly improve the performance of the model. Wordnet focuses on modeling relations between words instead of classification of words. The same level semantic class feature of two different words might be different in scale. For example, root hypernym of "people" is "entity.n.01" while root hypernym of "get" is "get.v.01", which might brings influence to the model.

Experimental result on English GEC dataset demonstrates that our proposed SG-GEC model could also benefit GEC task of other languages.

\section{More Case Studies}
\label{app:more_case}

We list five more cases in Table \ref{tab:more_case} to demonstrate effectiveness of our pipeline structure. BART model might easily miss grammatical error (Case 1, Case 2) or spelling error (Case 3, Case 4) because of mixing spelling error and grammatical error correction together. It might be misguided by erroneous token (Case 5). 

\begin{CJK}{UTF8}{gbsn}
\begin{table*}[t]
    \centering
    \begin{tabular}{lp{13cm}}
        \toprule[1pt]
         Type & Sample  \\
         \midrule
         \midrule
         SRC &  我认为空气污染是跟我们的生活密切的问题，所以一定要最优先解决，优其是像北京那样的大城市。\\
         TGT &  我认为空气污染是跟我们的生活密切\textcolor{blue}{相关}的问题，所以一定要最优先解决，\textcolor{red}{尤}其是像北京那样的大城市。\\
         BART & 我认为空气污染是跟我们的生活密切的问题，所以一定要最优先解决，\textcolor{red}{尤}其是像北京那样的大城市。\\
         SEC & 我认为空气污染是跟我们的生活密切的问题，所以一定要最优先解决，\textcolor{red}{尤}其是像北京那样的大城市。\\
         SG-GEC & 我认为空气污染是跟我们的生活密切\textcolor{blue}{相关}的问题，所以一定要最优先解决，\textcolor{red}{尤}其是像北京那样的大城市。\\
         Translation & I think air pollution is the problem that are closely related to our lives. Therefor it should be solved as a matter of top priority, especially for metropolis like Beijing. \\
         \midrule
         SRC &  人为了生存不管是干静的空气，污染的空气都要呼吸。\\
         TGT & 人为了生存不管是干\textcolor{red}{净}的空气，\textcolor{blue}{还是}污染的空气都要呼吸。\\
         BART & 人为了生存不管是干静的空气，\textcolor{blue}{还是}污染的空气都要呼吸。 \\
         SEC & 人为了生存不管是干\textcolor{red}{净}的空气，污染的空气都要呼吸。\\
         SG-GEC & 人为了生存不管是干\textcolor{red}{净}的空气，\textcolor{blue}{还是}污染的空气都要呼吸。 \\
         Translation &  In order to survive, human need to breathe air no matter it is fresh or polluted.\\
         \midrule
         SRC & 学校里的草场上还有一点的人来运动。 \\
         TGT & 学校里的\textcolor{red}{操}场上还有\textcolor{blue}{一些}人\textcolor{blue}{在}运动。\\
         BART & 学校里的草场上还有\textcolor{blue}{一些}人来运动。 \\
         SEC & 学校里的\textcolor{red}{操}场上还有一点的人来运动。\\
         SG-GEC &  学校里的\textcolor{red}{操}场上还有\textcolor{blue}{一些}人来运动。\\
         Translation & In the school playground, there are some people coming for doing exercise. \\
         \midrule
         SRC & 几个仙女来蟠桃圆摘桃时，告诉了孙悟空王母要做蟠桃盛会。\\
         TGT & 几个仙女来蟠桃\textcolor{red}{园}摘桃时，告诉了孙悟空王母要\textcolor{blue}{办}蟠桃盛会。\\
         BART & 几个仙女来蟠桃圆摘桃时，告诉了孙悟空王母要\textcolor{blue}{举办}蟠桃盛会。 \\
         SEC & 几个仙女来蟠桃\textcolor{red}{园}摘桃时，告诉了孙悟空王母要做蟠桃盛会。\\
         SG-GEC & 几个仙女来蟠桃\textcolor{red}{园}摘桃时，告诉了孙悟空王母要\textcolor{blue}{举办}蟠桃盛会。\\
         Translation & When fairies coming to peach orchard to pick peaches, they told Monkey King that the Queen Mother would hold the Peach Banquet. \\
         \midrule
         SRC & 这些地方征明中国灿烂的文化和历史。\\
         TGT & 这些地方\textcolor{red}{证}明\textcolor{blue}{了}中国灿烂的文化和历史。\\
         BART & 这些地方象征着中国灿烂的文化和历史。 \\
         SEC & 这些地方\textcolor{red}{证}明中国灿烂的文化和历史。\\
         SG-GEC & 这些地方\textcolor{red}{证}明\textcolor{blue}{了}中国灿烂的文化和历史。\\
         Translation & These places prove that China have splendid culture and history.\\
         \bottomrule[1pt]
    \end{tabular}
    \caption{Case study of our model. Red / blue color refers to correction of spelling / grammatical error.}
    \label{tab:more_case}
\end{table*}
\end{CJK}

\end{document}